\documentclass[]{interact}

\usepackage{algorithmic}
\usepackage{siunitx}
\usepackage{graphicx} 
\usepackage{capt-of}
\usepackage{amsmath}
\usepackage{amssymb}
\usepackage{multirow}
\usepackage{fancyhdr}
\newcommand{\quotes}[1]{``#1"}

\usepackage{epstopdf}
\usepackage[caption=false]{subfig}

\usepackage[natbibapa,nodoi]{apacite}
\theoremstyle{plain}

\theoremstyle{definition}

\theoremstyle{remark}

\begin{document}


\title{Towards Autoencoding Variational Inference for Aspect-based Opinion Summary}

\author{
\name{Tai Hoang\textsuperscript{a}\thanks{CONTACT Tai Hoang. E-mail: 13521093@gm.uit.edu.vn}, Huy Le\textsuperscript{a}\thanks{CONTACT Huy Le. E-mail: 13520360@gm.uit.edu.vn}, and Tho Quan\textsuperscript{b}\thanks{CONTACT Tho Quan. E-mail: qttho@hcmut.edu.vn}}
\affil{\textsuperscript{a}University of Information Technology, Thu Duc District, Ho Chi Minh City, Viet Nam; \\ \textsuperscript{b}Ho Chi Minh City University of Technology, District 10, Ho Chi Minh City, Viet Nam}
}


\maketitle

\begin{abstract}
 \emph{Aspect-based Opinion Summary} (AOS), consisting of \textit{aspect discovery} and \textit{sentiment classification} steps, has recently been emerging as one of the most crucial data mining tasks in e-commerce systems. Along this direction, the LDA-based model is considered as a notably suitable approach, since this model offers both topic modeling and sentiment classification. However, unlike traditional topic modeling, in the context of aspect discovery it is often required some initial \textit{seed words}, whose prior knowledge is not easy to be incorporated into LDA models. Moreover, LDA approaches rely on sampling methods, which need to load the whole corpus into memory, making them hardly scalable. 
In this research, we study an alternative approach for AOS problem, based on \emph{Autoencoding Variational Inference} (AVI). Firstly, we introduce the \emph{Autoencoding Variational Inference for Aspect Discovery} (AVIAD) model, which extends the previous work of \emph{Autoencoding Variational Inference for Topic Models} (AVITM) to embed prior knowledge of seed words. This work includes enhancement of the previous AVI architecture and also modification of the loss function. Ultimately, we present the \emph{Autoencoding Variational Inference for Joint Sentiment/Topic} (AVIJST) model. In this model, we substantially extend the AVI model to support the JST model, which performs topic modeling for corresponding sentiment. The experimental results show that our proposed models enjoy higher topic coherent, faster convergence time and better accuracy on sentiment classification, as compared to their LDA-based counterparts.

\end{abstract}


\section{Introduction}

\begin{figure}
\centering
\includegraphics[width=14cm,keepaspectratio]{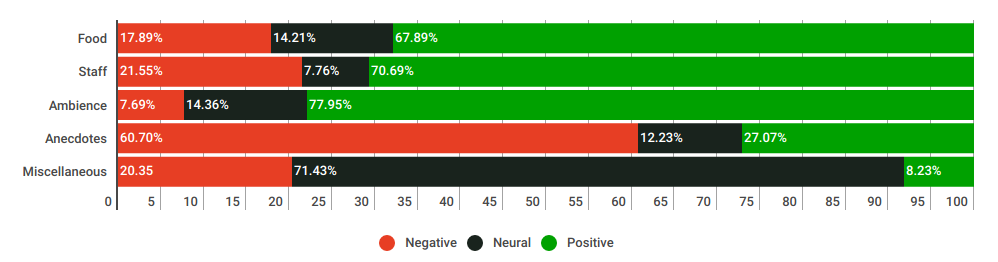}
\caption{Aspect-based Opinion Summary result of one product.}
\label{fig:intro}
\end{figure}

Recently, \emph{Aspect-based Opinion Summary} \cite{Hu2004-sc} has been introduced as an emerging data mining process in e-commerce systems. Generally, this task aims to extract \emph{aspects} from a product review and subsequently infer the sentiments of the review writer towards the extracted aspect. The result of an AOS task is illustrated as Fig.~\ref{fig:intro}. 
Thus, AOS consists of two major steps, known as \emph{aspect discovery} and \emph{aspect-based sentiment analysis}. For the first step, there are two major approaches. The first one relied on linguistic methods, such as using part-of-speech and dependency grammars analysis \cite{Qiu2011-nu} or using supervised methods \cite{Jin2009-wq}. However, this approach is likely able to detect only the \emph{explicit aspects}, e.g. the aspects which are referred explicitly in the context. 
For example, the review such as \emph{\quotes{The price of this restaurant is quite high}} can be inferred as a mention of the aspect \emph{price}, explicitly discussed in the text. However, for another review of \emph{\quotes{The foods here are not very affordable}}, the \emph{price} aspect is also implied, but implicitly. Thus, it is hard to be detected if one only fully relies on linguistic and supervised methods. 
The another approach for aspect discovery is based on \emph{Latent Dirichlet Allocation} (LDA) \cite{Blei2003-dv}, 
which is widely used for \emph{topic modeling} \cite{Zhao2010-bt}. In this approach, a \emph{topic} is modeled as a distribution of words in the given corpus, thus can be treated as a discovered aspect. For example, the \emph{price} aspect can be discovered as a distribution over some major words such that \emph{price, expensive, affordable, cheap}, etc. This approach is widely applied today to detect hidden topics in documents.
For the second step of aspect-based sentiment analysis, various works based on feature-extracted machine learning are proposed, e.g. \cite{Bespalov2011-ti}. Recently, many works on using deep learning for sentiment classification have also been report \cite{2018arXiv180107883Z}. 

However, in the context of topic discovery, perhaps the most remarkable work perhaps is the approach of \emph{Joint Sentiment/Topic} model (JST) \cite{Lin2009-dx} since this work extended the usage of LDA for topic modeling as a joint system allowing not only topics to be discovered but also sentiment words associated with the topics. Thus, it is very potential to completely solve the full AOS problem using LDA-based approach, as attempted in \cite{2015arXiv151109128W}.   

However, aspect discovery is not entirely the same process as topic modeling. As reported in \cite{Lu2011-wh}, the task of aspect discovery should require some initial words for the aspects, known as the \emph{seed words}. However, in LDA-based approaches, it is not easy to incorporate the prior knowledge of seed words to the topic modeling systems. Moreover, the LDA-based approaches rely on \emph{sampling methods}, e.g. \emph{Gibbs sampling} \cite{Blei2003-dv} 
to learn the parameters of the required Dirichlet distributions. This work requires the whole corpus to be loaded into memory for sampling, which incurs heavy computational cost. Then, in this study, we explore the application of \emph{Autoencoding Variational Inference For Topic Models} (AVITM) \cite{2017arXiv170301488S} as an alternative method of LDA for AOS. In AVITM, a deep neural network is integrated in \emph{variational autoencoder} \cite{DBLP:journals/corr/KingmaW13} with the technique of
\emph{reparameterization trick} 
to simulate the sampling work, which eventually learns the desired topic distribution as done by LDA.

Hence, the approach of AVI can achieve theoretically the same objective of hidden topic detection like LDA, but it can avoid the heavy cost of loading the whole corpus for sampling, since the input data can be gradually fed to the input layer of the deep neural network. Further, when the training data are enriched with new documents, the sampling process of LDA must be restarted from beginning, whereas in ATITM, the new documents can be incrementally trained in the neural network. Lastly, by unnormalizing the distribution of words with corresponding topics when training the neural networks, ATIVM can obtain more coherence on the generated topics. 

Urged by those advantages of the AVI-based approach, we consider further extending this direction in the theme of AOS. To be concrete, we consider using AVI to support aspect discovery, not only topic modelling. In addition, we also aim to introduce an AVI-based version of the JST model, which can also perform topic modeling and sentiment classification at the same time, meanwhile still enjoying the advantages offered by AVI as aforementioned. Thus, our research contributes on the two following novel modes. The first proposal is known as \emph{Autoencoding Variational Inference for Aspect Discovery (AVIAD)}, in which we extend the existing work of AVITM to support incorporating prior knowledge from a set of pre-defined seed words of aspects for better discovery performance. The second model is referred to as \emph{Autoencoding Variational Inference for Joint Sentiment/Topic} \emph{(AVIJST)}. This is our ultimate model, which can be considered as a counterpart of the JST model. However, since the autoencoders are used instead of sampling, AVIJST can easily be scalable. In addition, this model can return not only the sentiment/topic-word matrix $((S\times K) \times V)$, but also the sentiment-word matrix $(S \times V)$, which is useful in many practical situations. In addition, AVIJST can take into account the guidance from a small set of labeled data to achieve significant improvement on classification performance.
The rest of the paper is organized as follows. In Section~\ref{section:intro_avitm} we recall background knowledge of LDA and AVITM. In Section~\ref{section:aviad} and Section~\ref{section:intro_aviaos}, we present the models of AVIAD and AVIJST respectively. Section~\ref{section:exp} discusses our experimental results on some benchmark databases. Finally, Section~\ref{section:conclution} concludes the paper.

\section{Latent Dirichlet Allocation and Autoencoding Variational Inference Approaches for Topic Modeling} \label{section:intro_avitm}

In this section, we recall the technique of Autoencoding Variational Inference For Topic Model (AVITM) where autoencoder is adopted to play the role of Latent Dirichlet Allocation (LDA) for topic modeling.

\subsection{Latent Dirichlet Allocation and Joint Sentiment/Topic model} \label{section:tm_lda}

Given a large dataset of document, or \emph{corpus}, topic modeling is a unsupervised classification task that determines \textit{themes} (or \textit{topics}) in documents. In this context, a topic is treated as a distribution over a fixed vocabulary and a document can exhibit multiple topics (but typically not many). 
To fulfill this task, \textit{Latent Dirichlet Allocation} (LDA) is introduced as a generative process where each document is assumed to be generated by this process. Meanwhile, \textit{Joint Sentiment-Topic} (JST) model \cite{Lin2009-dx} is a generative model extended from the popular LDA model which is introduced to solve the problem of sentiment classification without prior labeled information. 
To generate a document,  the process randomly chooses a distribution over topics. Then, each word in the document is generated by randomly choosing a topic from the distribution over topics and then randomly choosing a word from the corresponding topic.

\begin{figure}
\centering
\subfloat[LDA]{%
\resizebox*{4.9cm}{!}{\includegraphics{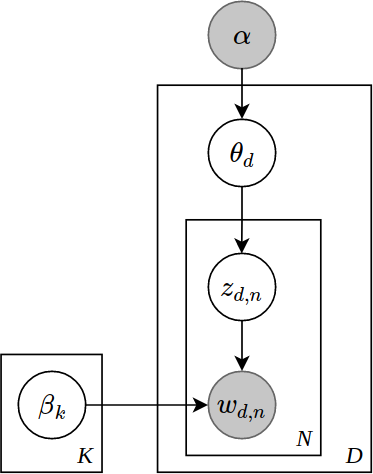}}
\label{fig:lda}
}\hspace{35pt}
\subfloat[JST]{%
\resizebox*{6.0cm}{!}{\includegraphics{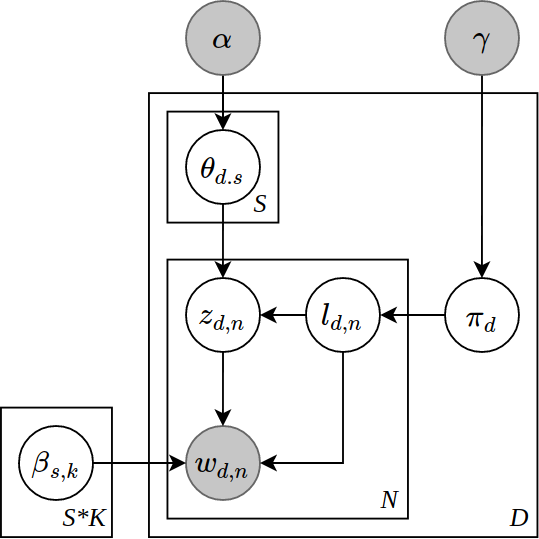}}
\label{fig:jst}
}
\caption{Probabilistic Graphical model of LDA and JST} \label{fig:graph_model}
\end{figure}

Formally, the LDA process and JST can be visualized as  graphical models given in Fig.~\ref{fig:lda} and Fig.~\ref{fig:jst} where
\begin{description}
    \item[$\bullet$] $\beta_{1:K}$ are the topic distribution and each $\beta_k$ is a distribution over the vocabulary correspondingly to topic $k$; 
    \item[$\bullet$] $\theta_{d}$ are the topic proportions for document $d$; 
    \item[$\bullet$] $z_{d,n}$ is the topic assignment for word $n$ in document $d$; 
    \item[$\bullet$] $w_{d,n}$ are the observed words for document $d$;
    \item[$\bullet$] $\alpha$ is the prior parameter of the respective Dirichlet distributions where $\theta_{d}$ is assumed.
\end{description}
A graphical model of JST is represented in Fig.~\ref{fig:jst}. Compared to LDA, JST has additionally the following component.
\begin{description}
  \item[$\bullet$] $\pi_{d}$ are the sentiment proportions for document $d$; 
  \item[$\bullet$] $l_{d,n}$ is the sentiment assignment for word $n$ in document $d$;
 \item[$\bullet$] $\alpha$ and $\gamma$ are the prior parameters of the respective Dirichlet distributions where $\theta_{d,s}$ and $\pi_{d}$ are assumed respectively.
\end{description}



Intuitively, the key idea behind the LDA process is that given a set of observed document $d$ over a vocabulary of $N$ words, we try to infer 2 sets of \emph{latent variables}, which are represented by document-topic distribution  and a topic-word distribution. Meanwhile, in JST, we try to infer 3 sets of latent variables, which is joint sentiment/topic-document distribution $\theta$,  joint sentiment/topic-word distribution $\beta$ as well as sentiment-document distribution $\pi$ given set of observed document $d$ over a vocabulary of $N$ words. 

Then, a document in the LDA process will be generated as

\begin{equation}
  p (w | \alpha, \beta) = \int_{\theta} \left(\prod_{n=1}^N \sum_{z_n=1}^k p(w_n | z_n, \beta) p(z_n | \theta) \right) p(\theta | \alpha)d\theta.
\end{equation}


Meanwhile, in the JST process, each document $d$ is generated from distribution:

\begin{equation}
\label{eq:jst}
\begin{split}
  p (w | \alpha, \beta, \gamma)={} \int & \prod_{s} p(\theta_s|\alpha) \int \bigg (  p(\pi| \gamma) \prod_{n} \sum_{s_n} p(s_n | \pi) \\
         & \sum_{z_n} p(z_n | s_n, \theta_s) p(w_n | z_n, s_n, \beta) d{\pi} \bigg ) d{\theta}.
\end{split}
\end{equation}

In LDA and JST approaches, those distributions are evaluated by sampling methods such as Gibbs sampling \cite{Lin2009-dx}. However, this sampling approach required the whole corpus to be loaded into memory, which is heavily costly. Moreover, the sampling approaches prevent concurrent processing and needed to be restated when there are changes in the dataset, making this direction hardly scalable.


\subsection{Variational Auto-Encoder} \label{section:vae}

\begin{figure}[ht]
\centering
\includegraphics[width=8.5cm, keepaspectratio]{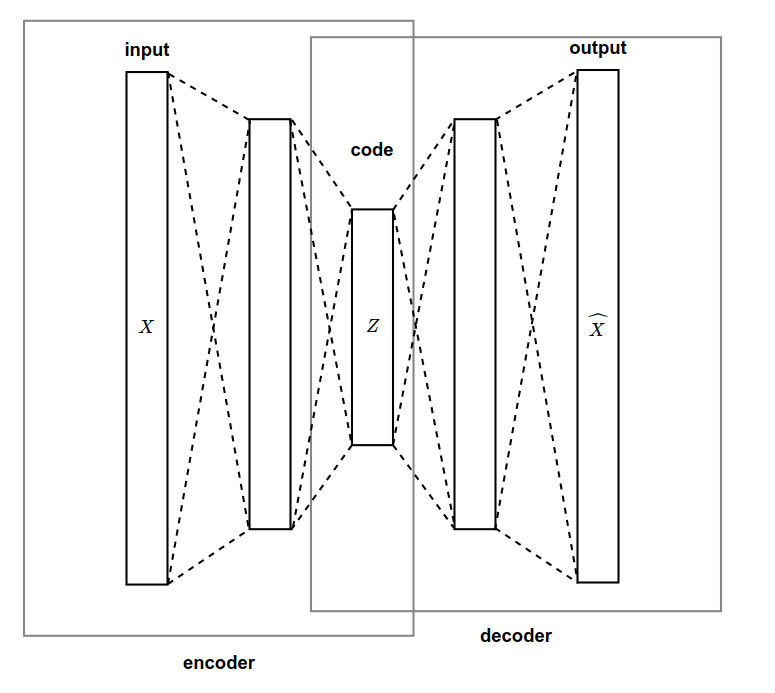}
\caption{A general autoencoder system}
\label{fig:ae}
\end{figure}

\emph{Autoencoder} (AE) \cite{Rumelhart:1986:LIR:104279.104293} 
is a neural network architecture which can be seen as a \emph{nonlinear} function (black-box) that includes two parts: encoder and decoder, as depicted in Fig.~\ref{fig:ae}. 

Given an input $x$ in a higher-dimensional space, the encoder maps $x$ into $z$ in a lower-dimensional space as $z = f(x)$. Meanwhile, the decoder subsequently maps $z$ into $\hat{x}$ as $\hat{x} = g(z)$ where $\hat{x}$ is in the same space as $x$. The loss function of the overall network will be calculated as $\mathcal{L}(x,\hat{x}) = \|x - \hat{x}\|^2$, whose aim is to make the decoder generate the $\quotes{same}$ input given. Once the network converges, the encoded $z$ will represent hidden features (or \textit{latent features}) discovered from the input space.

However, the latent space generated by the traditional Autoencoder process is generally concrete (not continuous), thus it can well generate latent features on the samples which were previously trained. However, when generating latent information for a new sample, this method may suffer from poor performance.




\begin{figure}[ht]
\centering
\includegraphics[width=14cm, keepaspectratio]{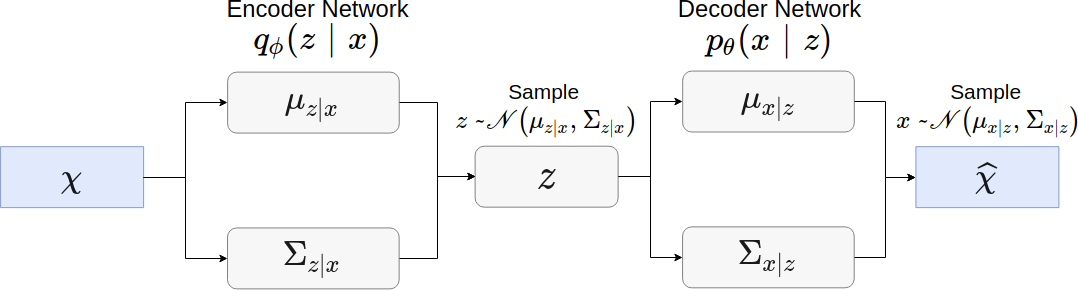}
\caption{Variational Auto Encoder model where observable variable $x$ and its correspondent latent $z$ are distributed on Gaussian distribution $\mathcal{N}(\mu_{x|z}, \Sigma_{x|z})$ and $\mathcal{N}(\mu_{z|x}, \Sigma_{z|x})$, respectively.}
\label{fig:vae}
\end{figure}

Variational Auto-Encoder (VAE) \cite{DBLP:journals/corr/KingmaW13} is an extension of AE, where the latent space $z$ will be learn a posterior probability $p(z|x)$, as depicted in Fig.~\ref{fig:vae}. Thus, the encoding-decoding process will be performed as follows. For an input datapoint $x_i$, the encoder will draw a latent variable $z$ by sampling based on the posterior probability $p(z|x)$. Then, the decoder will generate the output $\hat{x}$ by sampling based on the posterior probability $p(x|z)$. In other words, the encoder tries to learn $p(z|x)$, whereas the decoder tries to learn $p(x|z)$.

Based on Bayes theorem, $p(z|x)$ can be evaluated as
\begin{equation}
\centering
\label{eq:bayes}
  p(z|x) = \frac{p(x|z)p(z)}{p(x)}.
\end{equation}

However, since the prior distribution $p(x)$ is generally intractable, we will instead approximate the desired probability $p(z | x)$ by an approximated distribution $q_\lambda(z|x)$ where $\lambda$ is the \emph{variational parameter} corresponding to the distribution family to which $q$ belongs. In VAE, $q$ is often a Gaussian distribution, hence  for each data point $x_i$, we have $\lambda_{x_i} = \{\mu_{x_i}, \sigma_{x_i}^2\}$. Moreover, VAE adopts the \emph{amortize inference} \cite{2016arXiv161005735R} approach, in which all of data points will share (amortize) the same $\lambda$.

Thus, the encoder part of an VAE will consists of two fully connected modules, whose roles are to respectively learn the parameter $\mu$ and $\sigma$, as presented in Fig.~\ref{fig:vae}. For an input vector $x$, a latent vector $z$ will be sampled based on the currently learned value of $\mu$ and $\sigma$. The decoder will do the similar thing to reconstruct the output $\hat{x}$.

The goal of the learning process is that we try to make the variational distribution $q_\lambda(z|x)$ as \quotes{similar} as possible to the $p(z|x)$. To measure such similarity, one can use the Kullback-Leibler divergence \cite{Cover:1991:EIT:129837}, which measures the information lost when using $q$ to approximate $p$:
\begin{equation}
\label{eq:KL_approx}
\begin{split}
  D_{KL}(q_\lambda(z | x)\|p(z|x)) = & \ \mathbb{E}_q\left[\log q_\lambda(z|x)\right] \\
  & - \mathbb{E}_q[\log p(x,z)] + \log p(x).
\end{split}
\end{equation}

Our goal is to find the variational parameters that minimize this divergence, or $q^*_\lambda(z|x) = \arg \min_\lambda  D_{KL}\left(q_\lambda(z | x)\|p(z | x) \right)$. From \ref{eq:KL_approx}, we have
\begin{equation}
\label{eq:elbo_0}
  \text{ELBO}(\lambda) = \mathbb{E}_q[\log p(x,z)] - \mathbb{E}_q\left[\log q_\lambda(z|x)\right].
\end{equation}
and
\begin{equation}
\label{eq:elbo_1}
  \log p(x) = \text{ELBO}(\lambda) + D_{KL}(q_\lambda(z | x)\|p(z|x)).
\end{equation}

where ELBO stands for \emph{Evidence Lower Bound}. It is due to the fact that the Kullback-Leibler divergence is always greater than or equal to zero, by Jensen\textsc{\char13}s inequality \cite{Cover:1991:EIT:129837}. Hence, instead of minimizing the Kullback-Leibler divergence in \ref{eq:elbo_1}, one can equally maximize $\text{ELBO}(\lambda)$.

When deployed in a VAE, $\text{ELBO}(\lambda)$ can be computed as ELBO of all data points. ELBO of a single datapoint can be expressed as
\begin{equation}
\label{eq:elbo_2}
  \text{ELBO}_i(\lambda) = \mathbb{E}_{q_\lambda(z|x_i)}[\log p(x_i|z)] - D_{KL}(q_\lambda(z | x_i)\|p(z)).
\end{equation}
where $p(z)$ is adopted as the normal distribution $\mathcal{N}(0,1)$.

Thus, let $\theta$ and $\phi$ be the weights and biases of the encoder and decoder of the VAE, Equation \ref{eq:elbo_2} can be regarded as the loss function of the VAE as
\begin{equation}
\label{eq:elbo}
  \mathcal{L} = \underbrace{E_{q_\phi(z | x)}\overbrace{\left [\log p_\theta(x | z) \right ] }^\text{decoder}}_\text{reconstruction loss} - \underbrace{D_{KL}(\overbrace{q_\phi(z | x)}^\text{encoder}\|\overbrace{p_\theta(z)}^\text{prior})}_\text{recognition loss}. \\
\end{equation}

where the \emph{reconstruction loss} measure the error occurring when the VAE reconstruct the output from the input, meanwhile the \emph{recognition loss} measure the error occurring when generating the latent variable, which play the role of regularization of this loss function.

The encoder networks usually simulates Gaussian distributions, so the recognition loss has nice closed-form solution. On the other hand,  the reconstruction loss can be estimated by using Monte-Carlo sampling. However, sampling method is generally indifferentiable and thus cannot be back-propagated in a \emph{NN} system. Thus, in  \cite{DBLP:journals/corr/KingmaW13}, the \emph{reparameterization trick} is proposed, which replaced the sampling step $z \sim p(z | x) = \mathcal{N}(\mu,\sigma^2)$ in the training process by $z = \mu + \sigma\epsilon$ where $\epsilon$ is sampled from the trivial distribution $\mathcal{N}(0,1)$, which is differentiable.

\subsection{VAE for topic modeling} \label{section:avitm}

From the original probability used by LDA, one can  use the \emph{collapsing z\textsc{\char13}s} technique \cite{2017arXiv170301488S} to reduce the number of distributions that we need to compute the approximation as 

\begin{equation}
\label{eq:avitm_decoder}
\begin{split}
  p (w | \alpha, \beta) & = \int_{\theta} \left(\prod_{n=1}^N \sum_{z_n=1}^k p(w_n | z_n, \beta) p(z_n | \theta) \right) p(\theta | \alpha)d\theta \\
  & = \mathbb{E}_{p(\theta | \alpha)} \left[ \prod_{n=1}^N p(w_n\mid\theta, \beta) \right].
\end{split}
\end{equation}

where $p (w | \alpha, \beta) = \text{Multinomial} \left(1, \sum_{k} \theta_k \beta_k \right )$. \\

Hence, one only needs to evaluate the distributions of $\theta$ and $\beta$.  In \cite{2017arXiv170301488S}, an approach using VAE for topic modeling has been introduced to replace the old approach of LDA.  VAE uses autoencoder to learn the distributions of $\theta$ and $\beta$. However, as LDA uses Dirichlet distribution, meanwhile VAE is intended to learn Gaussian distributions as previously discussed. To solve this, \emph{Laplace approximation} \cite{2017arXiv170301488S} is applied. Basically, a Dirichlet prior distribution $p(\theta | \alpha)$ with parameters $\alpha$ (for $K$ topics) will be approximated as $\mathcal{N}(\mu_1, \sigma_1)$  where $\mu_1$ and $\sigma_1$ are evaluated as below.


\begin{equation}
    \begin{split}
  \mu_{1k} &=\log \alpha_k - \frac{1}{K} \sum_{i}^K{\log\alpha_i}. \\
  \Sigma_{1kk} &= \frac{1}{\alpha_{k}} \left(1 - \frac{2}{K} \right) + \frac{1}{K^2}\sum_{i}^K{\frac{1}{\alpha_i}}.
    \end{split}
  \label{eq:laplace-approx}
\end{equation}

Thus, we can compute the recognition loss by evaluating the closed-form of KL divergence between two Gaussian distributions. On the other hand, the reconstruction loss is evaluated by compute the \emph{probability density function} of distribution $p (w | \alpha, \beta) = \text{Multinomial} \left ( 1, \sum_{k} {\theta_k \beta_k} \right )$. 
Therefore, the final loss function is computed as follows.

\begin{equation}
    \begin{split}
        \mathcal{L}(\Theta) & =  \sum_{d=1}^D 
        \bigg [  \mathbb{E}_{\epsilon \sim \mathcal{N}(0,I)}  \left [ w_d^T \log \left (\sigma(\beta)\sigma(\mu_0 + \Sigma_0^{1/2} \epsilon) \right ) \right ] \\ 
         & - \left ( \frac{1}{2} \left \{ tr(\Sigma_1^{-1}\Sigma_0) + (\mu_1 - \mu_0)^T\Sigma_1^{-1}(\mu_1 - \mu_0) - K + \log\frac{|\Sigma_1|}{|\Sigma_0|} \right \} \right ) \bigg ].
    \end{split}
    \label{eq:avitm_loss}
\end{equation}


\section{Autoencoding Variational Inference for Aspect Discovery} \label{section:aviad}

As previously discussed, \emph{aspect discovery} \cite{Liu-_Proceedings_of_the_52nd_annual_meeting_of_the2014-sz} is a problem similar to topic modeling. Instead of discovering topics, one tries to discover \emph{aspects} of \emph{concepts} mentioned in a document. For example, when analyzing reviews of restaurants, the aspects that can be mentioned may include \emph{food}, \emph{service}, \emph{price}, etc.

However, the key difference between topic modelling and aspect discovery is that the latter normally requires seed words \cite{Lu2011-wh}. Based on those seed words, other aspect-related terms are further discovered. In \cite{Lu2011-wh}, the discovery is done by  incorporating seed word information via prior distribution of $\beta$ distribution into \emph{Gibbs sampling} training process.

\begin{table}[ht]
\begin{minipage}[b]{0.45\linewidth}
    \centering
    \includegraphics[width=50mm,height=30mm,keepaspectratio]{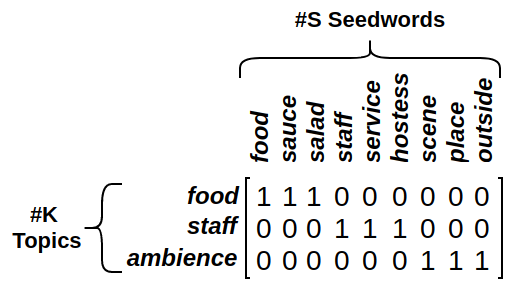}
    \captionof{figure}{Prior distribution matrix.}
    \label{fig:seed-matrix}
\end{minipage}\hfill
\begin{minipage}[b]{0.55\linewidth}
    \setlength{\tabcolsep}{4pt}
    \tbl{Discovered aspects.}
    {\begin{tabular}{lc} \toprule
     Aspect & Discovered Terms \\ \midrule
     Food & \textbf{sauce}, \textbf{salad}, cheese, onion, crab. \\
     Staff & \textbf{service}, \textbf{staff}, rude, \textbf{hostess}, waiter. \\
     Ambience & \textbf{scene}, \textbf{place}, wall, decorate, romantic. \\ \bottomrule \\
    \end{tabular}}
    \tabnote{\textbf{bold text} indicates seed words.}
    \label{table:aspect_discovered_by_gamma}
\end{minipage}
\end{table}

\begin{figure}[ht]
\centering
\includegraphics[width=13cm, keepaspectratio]{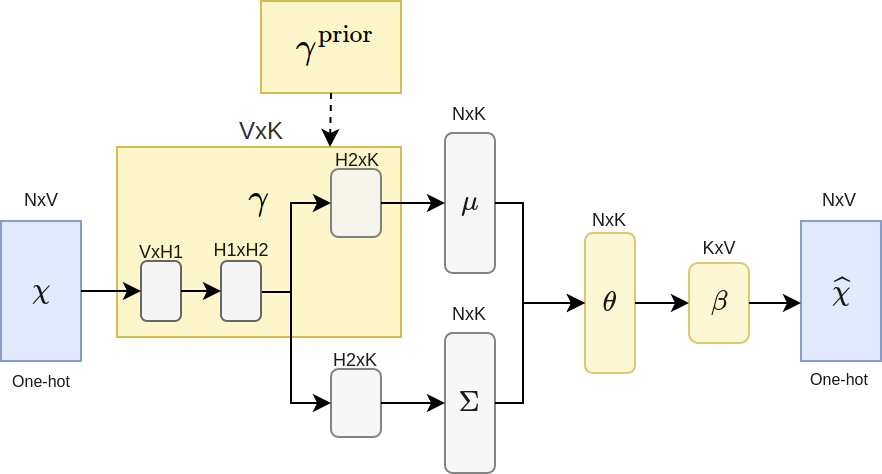}
\caption{AutoEncoding Variational Inference for Aspect Discovery. As illustrated, the yellow block $\theta$ and $\beta$ is corresponded to the document-topic and topic-word distributions which is described in Fig.~\ref{fig:lda}, respectively. Meanwhile, $\gamma$ and $\gamma^{\text{prior}}$ are additional blocks which play an important role in the aspect discovery task.}
\label{fig:m1}
\end{figure}

On the other hand,  in \emph{AVITM} \cite{2017arXiv170301488S}, the authors use non-smooth version \emph{LDA} illustrated in Fig.~\ref{fig:lda}, where $\beta$ has no prior distribution. Therefore, for \emph{VAE} direction, we propose to modify the loss function to reflect the prior knowledge conveyed by the seed words. It is realized by our proposed \emph{Autoencoding Variational Inference  for Aspect Discovery} (AVIAD) model, as presented in Fig.~\ref{fig:m1}. 

The goal of this model is also to retrieve the topic-word distribution $\beta$, like the original model described in Sect.~\ref{section:avitm}. However, in this model, we also embed the prior knowledge of seed words into the network structure. For example, 
the prior distribution $\gamma^{prior}$ of the given seed words will be represented as the matrix represented in Fig.~\ref{fig:seed-matrix}

The idea behind this distribution matrix is that we want to \quotes{force}, for instance, the seed word \emph{salad} to belong to the aspect \emph{Food}, which is represented as the first row in the matrix. In our AVIAD model, this prior distribution $\gamma^{prior}$ is given as yellow block in Fig.~\ref{fig:m1}. To incorporate this distribution in our training process, we introduce new loss function. 

\begin{equation}
  \begin{split}
  \mathcal{L}(\Theta) = \mathbb{E}_{q_{\phi}(\theta|w)} \left [ \log p (w | \theta, \beta) \right ] & - D_{KL} (q_{\phi} (\theta | w) \| p(\theta)) \\
  & - \lambda \sum_{n \in S} {\| \sigma(\gamma_n) - \gamma_n^{prior} \|^2}.
  \end{split}
  \label{eq:m1-loss}
\end{equation}

One can see that we have modified the ELBO in (\ref{eq:elbo}) by introducing the new term of (\ref{eq:m1-loss}), where each $\gamma_n$ is a topic distribution for each word $n$ that existed in set $S$ corresponding to document $d$. Thus, this square loss term ${|| \sigma(\gamma_n) - \gamma_n^\text{prior} ||^2}
\label{eq:loss-m1}$ will make the network try to produce the distribution $\gamma_n$ as  \emph{similar} to the prior distribution $\gamma^\text{prior}$ of seed words as possible. As a result, not only the predefined seed words are distributed to corresponding aspects, but also other similar words are also discovered in those aspects, as illustrated in Table~\ref{table:aspect_discovered_by_gamma}. 
\begin{figure}[ht]
\centering
\includegraphics[width=11.5cm, keepaspectratio]{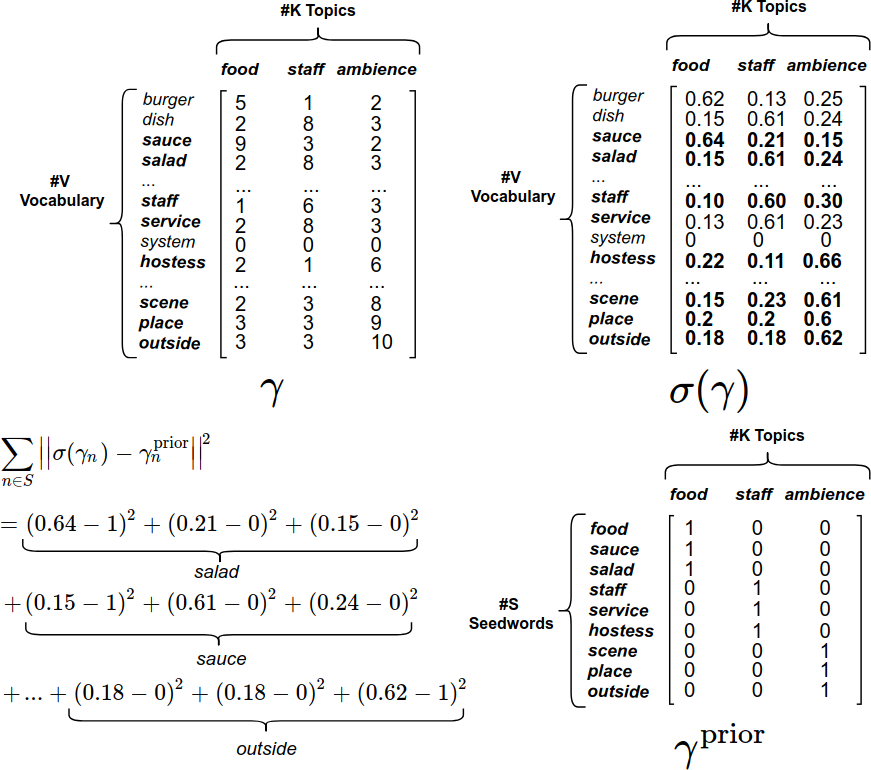}
\caption{Example of constructing the square loss term. Firstly, every rows in the $\gamma$ matrix is normalized via softmax function. After that, a submatrix is constructed by choosing only set of words in the normalized matrix $\gamma$ such that this word must also exists in a given $\gamma^{\text{prior}}$ matrix. Finally, the prior loss is computed via the \emph{Euclidean distance} between this submatrix and $\gamma^{\text{prior}}$. }
\label{fig:m1explain}
\end{figure}

For example, assume at the $k^{th}$ iteration of training process, the learning matrix $\gamma$, given as illustrated in Fig.~\ref{fig:m1explain}, must be normalized by applying softmax function $\sigma$. Then, by minimizing the \emph{Euclidean} distance between it and the prior matrix $\gamma^\text{prior}$ in concurrently with the other two term in Equation \ref{eq:m1-loss}, not only the injected word such as {\emph{sauce}, \emph{salad}}, but \emph{onion, cheese} will also be converged to the true aspect. 

  
\section{AutoEncoding Variational Inference  for Joint Sentiment/Topic} \label{section:intro_aviaos}

\subsection{The Proposed Model of AVIJST} \label{section:proprose_aviaos} \label{section:aviaos}

In this section, we discuss our proposed ultimate model of \emph{Autoencoding Variational Inference for Aspect-based Joint Sentiment/Topic (AVIJST)}.
Instead of training the \emph{JST} model using \emph{Gibbs sampling}, we want to take the advantage of \emph{Variational Autoencoder} method which is fast and scalable on large dataset to this joint sentiment/topic model. First, inspired from the \emph{collapsing z\textsc{\char13}s} technique \cite{2017arXiv170301488S}, we collapse both set of $z$ and $s$ variables
. Thus, we only have to sample from $\theta$ and $\pi$ only:

\begin{equation}
\begin{split}
     p (w | \alpha, \beta, \gamma) &= \int \prod_{s} p(\theta_s | \alpha) \int p(\pi | \gamma) \prod_{n} p(w_n | \pi, \theta_s, \beta) d{\pi}d{\theta} \\
     &= \mathbb{E}_{\prod_{s} p(\theta | \alpha) p(\pi | \gamma)} \left[ \prod_{s}p(w_n | \pi, \theta_{s}, \beta) \right].
\end{split}
\end{equation}

where $p (w_n | \pi, \theta, \beta) = \text{Multinomial} \left(1, \sum_{s} {\pi_s \sum_{k}{\theta_{sk}\beta_{sk}}} \right)$. 

\begin{figure}[ht]
\centering
\includegraphics[width=13cm, keepaspectratio]{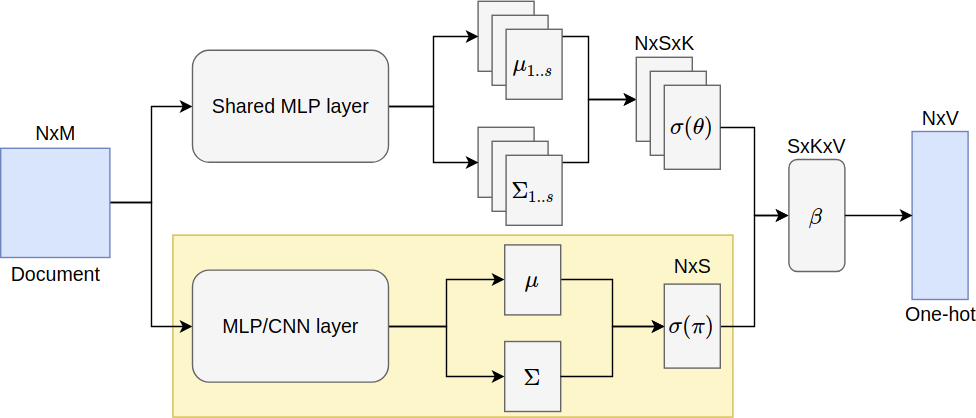}
\caption{AutoEncoding Variational Inference for Joint Sentiment/Topic Modeling where $\theta$, $\beta$ and $\pi$ are the corresponding latent variables in JST graphical model (Fig~.\ref{fig:jst}). Moreover, the yellow block which is used to mapping document $x$ to latent variable $\pi$ can be represented as any modern classification deep neural network.}
\label{fig:m2}
\end{figure}

In AVIJST, we no longer rely on the predefined set of seed words, since it is not easy to construct such a set with a large corpus. Instead, we observed that the distribution $\pi$ is trained through \textit{reparameterization trick} to reflect the sentiment of each document, so it can be seen as a discriminant function trained in supervised model. Motivated from this, 
we incorporate \emph{prior knowledge} by using \emph{labeled} information. That is, we use a (small) set of sentiment-labeled documents to guide the learning process. In the experiment, we can also treat our model as a semi-supervised model, which needs only a small set of labeled information for classification problem. The network structure of our AVIJST is given in Fig.~\ref{fig:m2}.

In our model, the classification network for $\pi$ distribution described in the yellow block in Fig.~\ref{fig:m2}, which is compatible with many kinds of neural networks. For simplicity, we only consider the \emph{Multi Layer Perceptron} (MLP) network and state-of-the-art \emph{Convolution Neural Network} (CNN) network combined with \emph{word embedding} (WE) layer. Besides, we use the same parameters (shared network) for all $\theta_{s}$ in hidden layers of encoder network instead of constructing $s$ different encoder networks. We want to remind the readers that the soft-max layer is applied for normalization purpose at the final layer of $\sigma(\theta)$ and $\sigma(\pi)$ variables. Then, our new lower-bound function 
is given as

\begin{equation} 
    \label{eq:m2-loss}
    \begin{split}
    \mathcal{L}  =&  \underbrace{\mathbb{E}_{\prod_sq(\theta_{s}|w)q(\pi|w)}\left[\log  \prod_n  p(w_n|\pi,\theta_{s},\beta)  \right  ]  }_\text{reconstruction  loss}  +  \underbrace{\mathbb{E}_{\hat{p}(\pi)}\left[-\log  q(\pi|w)\right  ]}_\text{classification  loss}  \\  
    &  -  \underbrace{  D_{KL}(q(\pi|w) \| p(\pi))  -  \sum_{s}  D_{KL}(q(\theta_s|w) \| p(\theta_s))}_\text{recognition loss}. 
    \end{split}
\end{equation}

As discussed in Sect.~\ref{section:avitm}, the reconstruction loss can be computed through Monte-Carlo sampling, while the recognition loss has nice closed-form since Laplace approximation is applied. In addition, to incorporate \emph{labeled information}, we integrate the classification loss in our loss function in (\ref{eq:m2-loss}) where $\hat{p}(\pi)$ is empirical distribution.

\subsection{Sentiment-word matrix} \label{section:sentiword_matrix}

One additional advantage of our proposed AVIJST is that we can generate the sentiment-word matrix from the learning results. Firstly, we observed that each word $w$ in a document is generated by $p (w | \pi, \theta, \beta) =$ Multinomial $ \left(1,\pi\theta\beta \right)$; where sentiment/topic-word distribution $\beta$ can be seen as a learning weight matrix in the decoder network which presented in Fig.~\ref{fig:m2}. 
Inspired from this, we want our model learn another sentiment-word distribution $\nu$, which show the top word for each sentiment orientation. 
We integrate $\nu$ in our model by combining it linearly to generate new generative distribution:

\begin{equation}
  \label{eq:m2-nu}
  p (w | \pi, \theta, \beta) = \text{Multinomial} (1, {\pi}{\theta}{\beta} + {\lambda}{\pi}{v}).
\end{equation}

In Equation (\ref{eq:m2-nu}), $\lambda$ is regularization weight. 
In experiments section, we will illustrate some top words resulted from the constructed sentiment-word matrix.

\section{Experiments} \label{section:exp}

\subsection{Datasets and Experimental Setup} \label{section:exp_datasets}
\subsubsection{Aspect Discovery} 
For evaluating the AVIAD model, the URSA\footnote{\url{http://spidr-ursa.rutgers.edu/datasets/}} restaurant dataset was used. It contains \num[group-separator={,}]{2066324} tokens and \num[group-separator={,}]{52624} documents in total. Documents in this dataset are marked with one or more labels from the standard label set $S$ = \{Food, Staff, Ambience, Price, Anecdote, Miscellaneous\}. In order to avoid ambiguity, we only consider sentences with single label from three standard aspects $S$ = \{Food, Staff, Ambience\}. Moreover, there are two different kind of dataset will be evaluated in the experiment of topic modeling (Section~\ref{section:exp_avids}) which are the \emph{imbalance} and \emph{balance} dataset. While only \num[group-separator={,}]{10000} sentences for each aspect is evaluated in \emph{balance} dataset, the \emph{imbalance} dataset contains \num[group-separator={,}]{62348}, \num[group-separator={,}]{23730} and \num[group-separator={,}]{13385} sentences in Food, Staff and Ambience aspect, respectively. Finally, due to the stability of the \emph{balance} dataset, it will also be used in the evaluation of supervised performance in Section~\ref{section:aviad_supervised}.

We set number of topics $K = 3$ corresponding to the number of chosen aspects, and Dirichlet parameter $\alpha = 0.1$ in our AVIAD model. 
In this experiments, we compare our method with the Weak supervised LDA (WLDA) \cite{Lu2011-wh}, which incorporates seed word information via prior distribution of latent variable $\beta$ in LDA model. Furthermore, the set of seed words which will be fed into both models is described in Table \ref{table:m1-seedword}.

\begin{table}[h!]
\tbl{AVIAD and WLDA seedwords.}
{\begin{tabular}{lc} \toprule
 Aspect & Seed words \\ \midrule
 \multirow{1}{*}{Food} 
 & food sauce shrimp cheese potato fry tomato \\ 
 \midrule
 \multirow{1}{*}{Staff} 
 & staff service friendly rude hostess waiter bartender waitress \\ 
 \midrule
 \multirow{1}{*}{Ambience} 
 & atmosphere scene place table outside outdoor ambiance \\
 \bottomrule
\end{tabular}}
\label{table:m1-seedword}
\end{table}

\subsubsection{Sentiment Classification}

Regarding AVIJST model, we use two different data sets which are Large Movie Review Dataset (IMDB)\footnote{\url{http://ai.stanford.edu/~amaas/data/sentiment/}} 
and Yelp restaurant \footnote{\url{https://www.yelp.com/dataset/challenge}}. The statistical information of these datasets are described in Table \ref{table:imdb}. Similar to the restaurant dataset, in the experiment of unsupervised performance, the sentiment datasets will also be devived into two subsets with different size, which are \{10k, 25k\} and \{20k, 200k\} for IMDB and Yelp, respectively. Moreover, due to the requirement of prior information, the subjectivity word list MPQA\footnote{\url{http://www.cs.pitt.edu/mpqa/}} is also used in the re-implementation of JST model.

In the experimental setup, since we want to avoid the collapsing problem which was proposed by \cite{2017arXiv170301488S}, the learning rate is set very high at $0.001$ for both AVIAD and AVIJST. In addition, in AVIJST, we also set the classification learning rate to $0.005$, due to the collapsing also occurred when training the discriminant distribution $\pi$.

\begin{table}[h!]
\tbl{Statistics for IMDB and Yelp datasets.}
{\begin{tabular}{lccccc} \toprule
 Dataset & \#labeled & \#unlabeled & \#testset & \#tokens & average length \\ \midrule
 IMDB & 25,000 & 50,000 & 25,000 & 11,737,946 & 235\\
 Yelp & 200,000 & 257,156 & 51,432 & 14,799,168 & 58\\
 \bottomrule
\end{tabular}}
\label{table:imdb}
\end{table}

\subsection{Experimental results of topic modeling} \label{section:exp_avids}

For topic modeling performance, we compare our proposed models of AVIAD and AVIJST with their LDA-based counterparts, i.e. the \emph{ Weakly supervised LDA} (WLDA) \cite{Lu2011-wh} and JST model with Gibbs sampling \cite{Lin2009-dx}, respectively.  In this experiment, we adopt the normalized point-wise mutual information (NPMI) \cite{Bouma-_Proceedings_of_GSCL2009-sb} as the main metrics to evaluate the qualitative of topic/aspects discovered by our model. The result has been proven that the set of word in discovered topic closely matches the human judgment in \cite{Lau2014-bx}. 



\begin{figure}[ht]
\centering
\resizebox*{14.0cm}{!}{\includegraphics{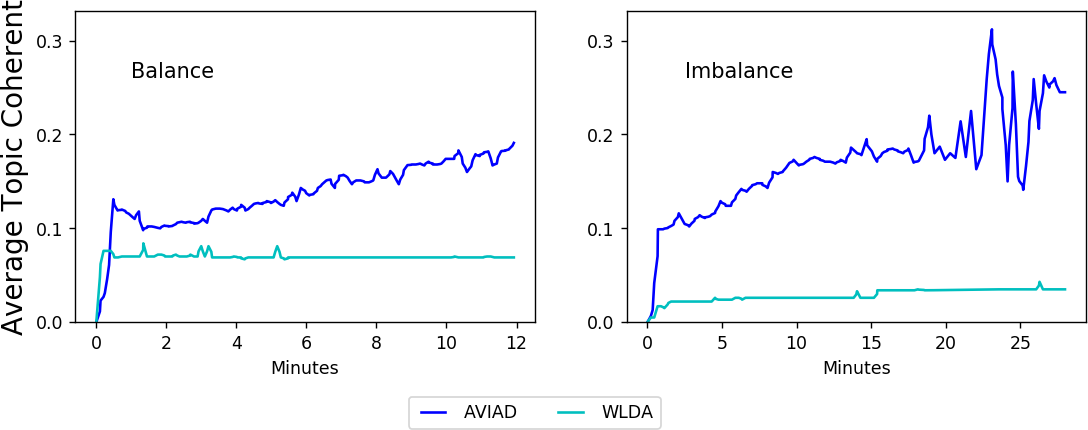}}
\caption{Aspect Discovery performance on restaurant dataset}
\label{fig:co_M1}
\end{figure}

\begin{figure}[ht]

\centering
\resizebox*{14.0cm}{!}{\includegraphics{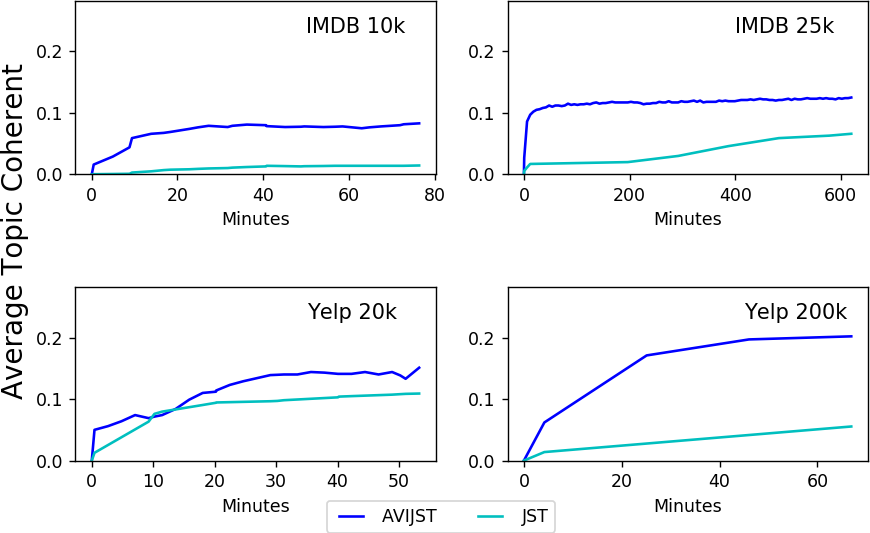}}
\caption{Joint Sentiment/Topic Modeling performance based on average topic coherent} \label{fig:co_M2}
\end{figure}

Fig.~\ref{fig:co_M1} and Fig.~\ref{fig:co_M2} show that AVIAD and AVIJST outperforms both the traditional WLDA and JST model where the average coherent values of our proposed models converge much faster than its counterpart models. Besides, we also report the top words discovered by AVIAD and AVIJST in Table \ref{table:aviad_topwords} and Table \ref{table:aviaos_jst}, \ref{table:aviaos_jst_yelp}, respectively. 

\begin{table}[h!]
\tbl{Topics extracted by AVIAD and WLDA.}
{\begin{tabular}{lcc} \toprule
 Model & Score & Top words \\ \midrule
 \multirow{3}{*}{AVIAD} 
 & 0.16 & scallop saute shrimp broccoli spinach tuna bean \\
 & 0.17 & apology bill refill behavior phone busboy manage \\
 & 0.28 & lit chandelier banquet dimly paint wooden mirror \\
 \midrule
 \multirow{3}{*}{WLDA} 
 & 0.09 & dish like pork good fresh taste menu \\
 & 0.06 & us food wait order time place even \\
 & 0.10 & great restaurant room nice like bar music \\
 \bottomrule
\end{tabular}}
\label{table:aviad_topwords}
\end{table}

\begin{table}[h!]
\tbl{IMDB Topics extracted by AVIJST and JST.}
{\begin{tabular}{lccccc} \toprule 
 \multicolumn{3}{c}{Positive} & \multicolumn{3}{c}{Negative} \\ \cmidrule{1-6}
 \multicolumn{6}{c}{\textbf{AVIJST}} \\ 
 \midrule
 Wrestle (0.31) & Starwar (0.30) & Kungfu (0 .29) & Horror (0.20) & War (0.22) & Novel (0.27) \\ 
 \midrule
 wwe & leia & hong & lugosi & germans & austen \\
 michaels & ewoks & cheung & bela & war & zelah \\
 undertaker & jedi & kung & karloff & soldiers & eyre \\
 hogan & darth & sammo & dracula & german & rochester \\
 hulk & vader & tsui & serum & hitler & novel \\
 cena & soldiers & shaolin & undead & soldier & novels \\ 
 \midrule
 \multicolumn{6}{c}{\textbf{JST}} \\ 
 \midrule
 Music (0.14) & Shows (0.11) & Kungfu (0.21) & Horror (0.13) & War (0.17) & Novel (0.15) \\ \midrule
 music & show & action & zombie & war & book \\
 musical & series & fight & horror & soldiers & the \\
 dance & the & japanese & zombies & military & read \\
 songs & episode & martial & gore & army & novel \\
 song & tv & fu & blood & soldier & books \\
 singing & episodes & kong & dead & battle & adaptation \\
 \bottomrule
\end{tabular}}
\label{table:aviaos_jst}
\end{table}

\begin{table}[h!]
\tbl{Yelp Topics extracted by AVIJST and JST.}
{\begin{tabular}{lccccc} \toprule
 \multicolumn{3}{c}{Positive} & \multicolumn{3}{c}{Negative} \\ \cmidrule{1-6}
 \multicolumn{6}{c}{\textbf{AVIJST}} \\ 
 \midrule
 Coffee (0.20) & Beer (0.26) & Thai food (0.36) & Staff (0.16) & Ambience (0.15) & Japanese food (0.34) \\ 
 \midrule
 coffee     & beers & pad 	    & rude 	        & reservation   & sushi \\
 latte 	    & beer 	& thai 	    & disrespectful & reservations  & sashimi \\
 espresso 	& brews & panang    & attitude  	& waited	    & nigiri \\
 baristas   & craft & curry     & unprofessional& closed 	    & rolls \\
 wifi 	    & tap 	& kha	    & filthy 	    & pm 		    & ayce \\
 study 	    & ipa   & thailand 	& yelled 	    & phone 	    & tempura \\
 \midrule
 \multicolumn{6}{c}{\textbf{JST}} \\ 
 \midrule
 Coffee (0.07) & Beer (0.13) & Japanese food (0.20) & Staff (0.08) & Ambience (0.13) & Thai food (0.19) \\ 
 \midrule
 coffee & beer      & sushi & staff     & wait          & thai \\
 shop   & bar       & roll  & rude      & table         & curry \\
 cafe   & selection & rolls & customers & seated        & pad \\
 local  & beers     & fish  & people    & pm            & chicken \\
 store  & tap       & fresh & customer  & night         & spicy \\
 latte  & pub       & tuna  & manager   & reservation   & rice \\
 \bottomrule
\end{tabular}}
\label{table:aviaos_jst_yelp}
\end{table}

Interestingly, Table~\ref{table:aviaos_jst} and \ref{table:aviaos_jst_yelp} shows that our AVIJST not only can extract the higher coherent topic, but also  discover more specific top words for each topic than the JST model. For example, in IMDB dataset, the words \quotes{shaolin}, which is a famous styles of Chinese material arts as well as \quotes{sammo}, who is a well known actor of Chinese, are founded in topic \quotes{Kung fu}, whereas JST model only retrieved the general words like \quotes{action} and \quotes{fight}. Moreover, our AVIJST can also discover topics that JST can not find, which are \quotes{Starwar} (0.30) and \quotes{Wrestle} (0.31). Meanwhile, in Yelp dataset, due to the independent to the polarity of some common words in documents, words in topic such as Thai food and Japanese food which were founded by AVIJST are inconsistent with JST model in term of polarity.
Furthermore, we also report the top words discovered for each sentiment orientation which mentioned in Sect.~\ref{section:sentiword_matrix} in Table \ref{table:m2-nu}.

\begin{table}[h!]
\tbl{Sentiment words discovered.}
{\begin{tabular}{lcc} \toprule
 Datasets & Polarity & Sentiment words \\ \midrule
 IMDB & Positive & {excellent, wonderful, underrated, superb, flawless} \\
  & Negative & {poorly, waste, worst, pointless, awful} \\ \midrule
 Yelp & Positive & {recommended, delicious, amazingly, underestimate, amicable} \\
  & Negative & {tasteless, flavorless, unimpressed, inedible, rude} \\ \bottomrule
\end{tabular}}
\label{table:m2-nu}
\end{table}


\subsection{Experimental results of classification performance} \label{section:exp_aviaos}





\subsubsection{Aspect Discovery} \label{section:aviad_supervised}

Although AVIAD and WLDA models were first proposed as an unsupervised model, the returned $\theta$ matrix can be treated as a classification model. Therefore, the classification performance of these models are also evaluated in Table \ref{table:m1_aspect_identification} via precision, recall and F1 metrics. 

In general, with the same set number of seedwords, our proposed AVIAD model outperformed WLDA in most case. Regarding the staff aspect, the amount of precision which is evaluated by WLDA model (0.662) is much lower than AVIAD (0.805) due the number of food aspects sentences is significantly larger than staff. Similarly, AVIAD recall is also 10\% greater than its counterpart model (0.793) on ambience aspect.


\begin{table}[h!]
\tbl{Aspect identification results.}
{\begin{tabular}{lccccc} \toprule
 & \multicolumn{4}{c}{Metrics} \\ \cmidrule{3-5}
 Aspect & Model & Precision & Recall & F1 \\ \midrule
 \multirow{2}{*}{Food} 
 & AVIAD & \textbf{0.948} & \textbf{0.806} & \textbf{0.870} \\
 & WLDA & 0.940 & 0.700 & 0.803 \\ \midrule
 \multirow{2}{*}{Staff} 
 & AVIAD & \textbf{0.805} & 0.887 & \textbf{0.842} \\ 
 & WLDA & 0.662 & \textbf{0.895} & 0.761 \\ \midrule
 \multirow{2}{*}{Ambience} 
 & AVIAD & 0.843 & \textbf{0.903} & \textbf{0.871} \\ 
 & WLDA & \textbf{0.844} & 0.793 & 0.817 \\ \bottomrule
\end{tabular}}
\label{table:m1_aspect_identification}
\end{table}

\subsubsection{Sentiment Classification}

To compare with our AVIJST, we build two others neural networks. The first one is multilayer perceptron classification network with two sequential fully connected hidden layers where input is Bag-of-word models for each document, called \emph{MLP}. For the second network, we construct \emph{CNN} network where each document is transformed into Word Embedding vector before feeding into Convolution Neural Network layers. These two networks are integrated in our AVIJST architecture under classification network with corresponding name \emph{AVIJST-MLP} and \emph{AVIJST-CNN} as presented in Fig.~\ref{fig:m2}. Furthermore, the model II semi supervised variational autoencoder (SSVAEII-MLP and SSVAEII-CNN) which was first proposed for semi-supervised problem in \cite{semi-vae} is also evaluated in this experiment.

\begin{table}[h!]
\tbl{Accuracy on test set for IMDB.}
{\begin{tabular}{lcccccccc} \toprule
 & \multicolumn{2}{l}{Labels} \\ \cmidrule{2-9}
 Model & 0 & 100 & 250 & 500 & 1k & 5k & 12k5 & Full \\ \midrule
 MLP & - & 68.5 & 77.1 & 80.5 & 82.5 & 84.8 & 86.1 & 87.0 \\
 CNN & - & 58.3 & 73.8 & 79.2 & 81.9 & 84.5 & 87.3 & 88.8 \\
 SSVAEII-MLP & - & 75.6 & 77.0 & 80.8 & 81.1 & 84.1 & 86.2 & 87.0 \\
 SSVAEII-CNN & - & 72.1 & 77.2 & 79.4 & 81.1 & 85.1 & 87.1 & 89.0 \\
 JST & 57.1 & - & - & - & - & - & - & - \\
 AVIJST-MLP & - & 68.2 & 72.7 & 79.9 & 81.2 & 85.9 & 88.2 & 89.7 \\
 AVIJST-CNN & - & \textbf{76.0} & \textbf{80.4} & \textbf{81.7} & \textbf{83.1} & \textbf{87.2} & \textbf{88.4} & \textbf{90.6} \\ \bottomrule
\end{tabular}}
\label{table:result-semi-m2-imdb}
\end{table}

\begin{table}[h!]
\tbl{Accuracy on test set for Yelp.}
{\begin{tabular}{lcccc} \toprule
 & \multicolumn{2}{l}{Labels} \\ \cmidrule{2-5}
 Model & 0 & 500 & 1000 & Full \\ \midrule
 MLP & - & 86.2 & 88.3 & 94.1 \\
 CNN & - & 85.1 & 87.0 & 94.5 \\
 SSVAEII-MLP & - & 87.2 & 87.8 & 94.5 \\
 SSVAEII-CNN & - & 84.5 & 86.0 & 94.5 \\
 JST & 78.0 & - & - & - \\
 AVIJST-MLP & - & 87.5 & \textbf{90.4} & 94.5 \\
 AVIJST-CNN & - & \textbf{88.4} & 89.8 & \textbf{94.9} \\ \bottomrule
\end{tabular}}
\label{table:result-semi-m2-yelp}
\end{table}

Due to the lack of knowledge from MPQA prior, the supervised performance of JST model is significantly lower than others. Meanwhile, the latent variables learned by our AVIJST model outperform the solitary classification network as well as the semi supervised VAE model in most cases which is shown in Table \ref{table:result-semi-m2-imdb} and \ref{table:result-semi-m2-yelp}. Especially, with using only 100 labeled documents over 25,000 in total on IMDB dataset, \emph{AVIJST-CNN} proved that the latent variables learned in our method can help the \emph{CNN} network achieve the highest accuracy (76.0 \%) among them, whereas the solitary \emph{CNN} showed high performance when given only a large number of labeled documents (half or full documents in the dataset in this case).


\section{Conclusion} \label{section:conclution}

In this paper, we study using Autoencoding Variational Inference approach for Aspect-based Opinion Mining, instead of the LDA-based approaches, widely known by the Joint Sentiment/Topic model. The motivation behind is that the deep neural networks of autoencoding allow us to avoid the heavy cost of sampling, enabling this approach scalable in parallel systems. This approach also enables us to take advantages of prior knowledge from seed words or small pre-labeled guiding sets to enjoy better performance.

As a result, we introduce two models of \emph{Autoencoding Variational Inference for Aspect Discovery} (AVIAD) and \emph{Autoencoding Variational Inference for Joint Sentiment /Topic} (AVIJST), which outperformed their LDA-based counterparts when experimented on benchmarking datasets. Especially, the AVIJST is designed flexibly which allows any neural-network-based classification method to be integrated in an end-to-end manner. Just for example, in this work we employed MLP and the state-of-the-art CNN deep networks for classification.

Even though our AVI-based approaches have been proven outperforming the LDA-based counterparts, we have still not fully solved the AOS problem by AVI. Thus, for the future work, we aim to a complete solution by investigating a neural network architecture allowing joint distributions of aspects, documents and sentiment to be represented and trained seamlessly.

\section*{Acknowledgements}

This research is funded by Vietnam National University HoChiMinh City (VNU-HCM) under grant number B2018-20-07.

\bibliographystyle{apacite}
\bibliography{interactapasample}

\appendix

\section{Generative Model}

In this section, we will show the generative process as well as its connection to Variational AutoEncoder in the decoder network of two aforementioned generative model LDA and JST.

\subsection{LDA} \label{section:appendix:lda}

Latent Dirichlet Allocation (LDA) assumes the following generative process for each document $d$ in a corpus $D$:

\begin{algorithmic}
  \FOR{document $d$ in corpus $D$}
  \STATE Choose $\theta_d \sim \text{Dirichlet}(\alpha) $
  \FOR{position $n$ in $d$}
    \STATE Choose a topic $z_{n} \sim \text{Multinomial}(1, \theta_d)$
    \STATE Choose a word $w_{n} \sim \text{Multinomial}(1, \beta_{z_{n}})$
  \ENDFOR
  \ENDFOR
\end{algorithmic}

Under this generative process, the marginal distribution of document $d$ is

\begin{equation}
\label{eq:lda_process}
 p (w | \alpha, \beta) = \int_{\theta} p(\theta | \alpha) \left(\prod_{n=1}^N \sum_{z_n=1}^K p(w_n | z_n, \beta) p(z_n | \theta) \right)d\theta.
\end{equation}

This equation \ref{eq:lda_process} can be seen as a model parameters form:

\begin{equation}
  p (w | \alpha, \beta) =  \frac{\Gamma(\sum_i\alpha_i)}{\prod_i\Gamma(\alpha_i)} \int_{\theta} \left( \prod_{i=1}^{K}\theta_i^{\alpha_i-1} \right ) \left(\prod_{n=1}^N \sum_{k=1}^K \prod_{j=1}^V (\theta_j \beta_{kj})^{w_n^j} \right )d\theta.
\end{equation}

Due to the one-hot encoding of $w_n^j$, thus, in VAE point of view, one can treat the generative distribution $p(w_n | \theta, \beta)$ as the decoder network.
\begin{equation}
    p(w_n | \theta, \beta) = \sum_{k=1}^K \prod_{j=1}^V (\theta_j \beta_{kj})^{w_n^j} = \left( \theta \beta \right)^{w_n}.
\end{equation}
where $\theta$ is the sampling matrix which is the output after using the reparameterization trick on the encoder network of $\mu_{\theta}$ and $\sigma_{\theta}$ while $\beta$ can be seen as a learning weight matrix of fully connected layer in the decoder network.

\subsection{JST} \label{section:appendix:jst}
A graphical model of JST is represented in Fig.~\ref{fig:jst}. Compared to LDA, JST has additionally the following component.
\begin{description}
 \item[$\bullet$] $\pi_{d}$ are the sentiment proportions for document $d$; 
 \item[$\bullet$] $l_{d,n}$ is the sentiment assignment for word $n$ in document $d$;
 \item[$\bullet$] $\alpha$ and $\gamma$ are the parameters of the respective Dirichlet distributions where $\theta_{d,s}$ and $\pi_{d}$ are assumed respectively.
\end{description}

Like LDA, each document $d$ is generated through a generative process

\begin{algorithmic}
  \FOR{document $d$ in corpus $D$}
  \STATE Choose $\pi_d \sim \text{Dirichlet}(\gamma) $
    \FOR{sentiment label $s$ under document $d$}
      \STATE Choose $\theta_{d,s} \sim \text{Dirichlet}(\alpha) $
    \ENDFOR
    \FOR{position $n$ in $d$}
     \STATE Choose a sentiment $l_{n} \sim \text{Multinomial}(1, \pi_d)$
     \STATE Choose a topic $z_{n} \sim \text{Multinomial}(1, \theta_{d,l_n})$
     \STATE Choose a word $w_n \sim \text{Multinomial}(1, \beta_{l_n,z_n})$
    \ENDFOR
  \ENDFOR
\end{algorithmic}

and its correspondent marginal distribution:

\begin{equation}
\label{eq:jst}
\begin{split}
 p (w | \alpha, \beta, \gamma)={} \int \prod_{s} p(\theta_s|\alpha) \int & p(\pi| \gamma) \prod_{n} \sum_{l_n} p(l_n | \pi) \\
        & \sum_{z_n} p(z_n | l_n, \theta_s) p(w_n | z_n, l_n, \beta) d{\pi}d{\theta}.
\end{split}
\end{equation}

where the reconstruction network can also be treated as a multiplication between three matrix $\theta$, $\beta$ and $\pi$.

\begin{equation}
    p(w_n | \theta, \beta, \pi) = \sum_{s=1}^S \sum_{k=1}^K \prod_{j=1}^V (\pi_s \theta_{sj} \beta_{skj})^{w_n^{j}} = \left(\pi \theta \beta \right)^{w_n}.
\end{equation}





\end{document}